\documentclass[letterpaper, 10 pt, conference]{ieeeconf}
\IEEEoverridecommandlockouts
\usepackage{preamble}

\title{\LARGE \bf
COFFAIL: A Dataset of Successful and Anomalous Robot Skill Executions in the Context of Coffee Preparation
}

\author{Alex Mitrevski$^{\mathsection,\dagger}$ and Ayush Salunke$^{\ddagger}$
\thanks{$^{*}$This work was supported by a starting research grant for Alex Mitrevski provided by Bonn-Rhein-Sieg University of Applied Sciences (project KEROL).} %
\thanks{$^{\dagger}$Alex Mitrevski is with the Division of Systems and Control, Chalmers University of Technology, Gothenburg, Sweden; {\tt\scriptsize alemitr@chalmers.se}} %
\thanks{$^{\dagger}$Ayush Salunke is with the Autonomous Systems Group, Bonn-Rhein-Sieg University of Applied Sciences, Sankt Augustin, Germany; {\tt\scriptsize ayush.salunke@smail.inf.h-brs.de}} %
\thanks{$^{\mathsection}$Corresponding author} %
}

\begin{document}

\maketitle
\thispagestyle{empty}
\pagestyle{empty}


\begin{abstract}
    In the context of robot learning for manipulation, curated datasets are an important resource for advancing the state of the art; however, available datasets typically only include successful executions or are focused on one particular type of skill.
    In this short paper, we briefly describe a dataset of various skills performed in the context of coffee preparation.
    The dataset, which we call COFFAIL, includes both successful and anomalous skill execution episodes collected with a physical robot in a kitchen environment, a couple of which are performed with bimanual manipulation.
    In addition to describing the data collection setup and the collected data, the paper illustrates the use of the data in COFFAIL to learn a robot policy using imitation learning.
\end{abstract}


    \section{INTRODUCTION}
    \label{sec:introduction}

    Robot skill execution datasets are an important resource that simplifies the investigation of robot learning and generalisation techniques, thereby enabling the development of flexible learning-based robot execution pipelines \cite{kroemer2021,firoozi2025}.
    The public availability of robot learning datasets has significantly increased in the last few years; however, large-scale learning datasets include only (or primarily) successful executions \cite{openx2024,bridge2023}.
    This makes such datasets suitable for learning policy models, but makes it challenging to use them for developing techniques for failure detection and recovery, which are also essential for guaranteeing robust execution.
    There are numerous datasets that include anomalous data, but they focus on either industrial applications \cite{reassemble2025,stowsuccess2025,armbench2023,screwdriving2021} or specific everyday skills, such as pouring \cite{imperfectpour2025}, object handover \cite{handover2024}, object placement \cite{placement2021}, or handle grasping \cite{handles2021}.
    There are a few datasets that include a larger variety of skills, including anomalies \cite{reflect2023,failure2021}; however, \cite{reflect2023} is primarily simulation-based (only a subset of the data is from a real robot), and both of these only include skills performed with a single manipulator.
    RoboMIND \cite{robomind2025} is another large dataset that includes data from numerous skills performed by four robot embodiments; RoboMIND includes anomalous executions, but not ones where the operation of the robot is deliberately obstructed, which are useful to consider for everyday robots.
    Most existing dataset also require an external camera setup, which limits a robot's operational environment significantly.

    In this short paper, we describe a dataset collected in the context of a domestic task, concretely that of coffee preparation; for this reason, we refer to the dataset as COFFAIL.
    COFFAIL is publicly available\footnote{Public dataset link: \url{https://doi.org/10.5281/zenodo.18212316}}$^{,}$\footnote{Dataset processing utilities: \url{https://github.com/KEROL-project/coffail-utils}}, and includes both successful and anomalous skill executions; most of the skills are performed by a single manipulator, but a few are bimanual skills.
    Below, we provide a brief description of the data collection setup and the collected data, and illustrate the use of the data for imitation learning.

    \section{DATASET DESCRIPTION}
    \label{sec:dataset_description}

\begin{figure}[t]
        \centering
        \includegraphics[width=\linewidth]{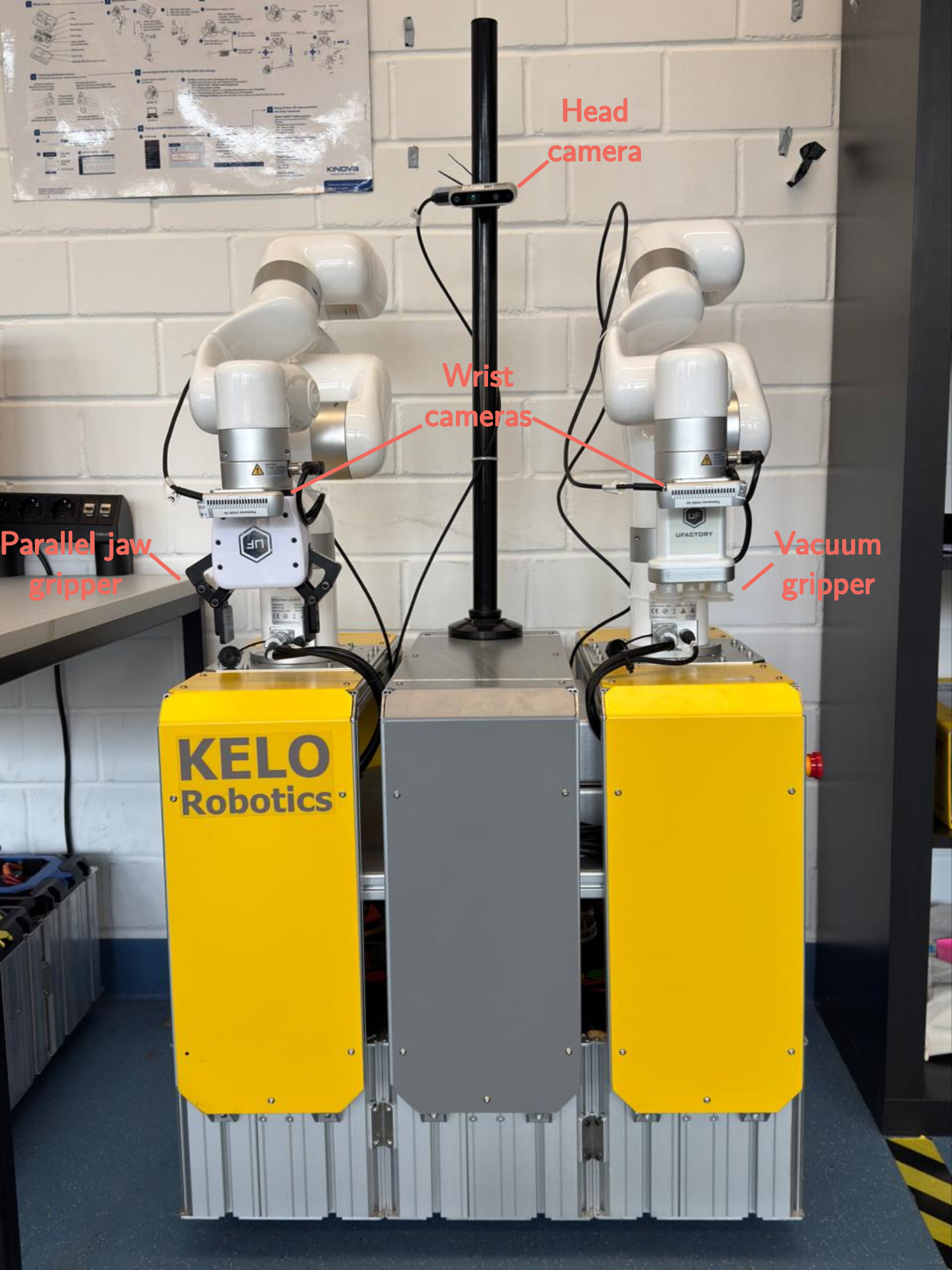}
        \caption{Jessie, the robot used for data collection}
        \label{fig:jessie}
    \end{figure}

    Our dataset was collected in a kitchen environment with a robot that is referred to as Jessie, shown in Fig. \ref{fig:jessie}.\footnote{\url{https://www.h-brs.de/en/a2s/robots}}
    Jessie is a bimanual mobile manipulator with a KELO ROBILE mobile base\footnote{\url{https://shop.kelo-robotics.com/product/standard-configuration/}} and two UFactory xArm 6 manipulators\footnote{\url{https://www.ufactory.cc/xarm-collaborative-robot/}}; one of the arms is equipped with a parallel jaw gripper, while the other arm has a vacuum gripper.
    The robot has three Intel RealSense RGB-D cameras (one head camera, which provides global scene view, and two wrist cameras, attached above the wrist of each arm), though we only collect the RGB images for our dataset.
    The robot's base is manually positioned in front of the location of interest before each execution episode, and does not move during execution.

    COFFAIL includes seven skills that are relevant in the context of domestic tasks.
    The skills, along with the number of collected episodes per skill, are summarised in Tab. \ref{tab:episode-counts}.
    \begin{table*}[t]
        \caption{A summary of the number of successful (\cmark) and failed (\xmark) episodes for each skill in the dataset}
        \label{tab:episode-counts}
        \begin{tabular}{M{0.08\linewidth} M{0.09\linewidth} M{0.09\linewidth} M{0.09\linewidth} M{0.09\linewidth} M{0.09\linewidth} M{0.09\linewidth} M{0.1\linewidth} M{0.07\linewidth}}
            \hline
            \cellcolor{gray!10!white}\textbf{Skill} & \cellcolor{gray!10!white}Cup pickup & \cellcolor{gray!10!white}Moving a cup & \cellcolor{gray!10!white}Pouring & \cellcolor{gray!10!white}Cup placing & \cellcolor{gray!10!white}Spoon pickup & \cellcolor{gray!10!white}Stirring & \cellcolor{gray!10!white}Placing a spoon in a sink & \cellcolor{gray!10!white}\textbf{Total} \\\hline
            \cellcolor{gray!10!white}\textbf{\cmark episodes}     & 15 & 12 & 12 & 10 & 10 & 10 & 10 & \textbf{79} \\\hline
            \cellcolor{gray!10!white}\textbf{\xmark \, episodes}  &  6 &  8 & 10 &  6 &  8 &  4 &  6 & \textbf{48} \\\hline
        \end{tabular}
    \end{table*}
    All episodes were collected by demonstration; most demonstrations were performed through hand-coded execution scripts that guarantee the desired behaviour, but some were collected using kinaesthetic teaching for increased variability.
    The anomalous episodes include various types of anomalies, such as missing objects, temporarily blocked cameras, collisions, or failed executions.
    The global scene was mostly static throughout the execution of each skill, but the positions of the objects and the starting position of the robot were varied between episodes.

\begin{figure*}[t]
        \begin{subfigure}{0.195\linewidth}
            \includegraphics[width=\linewidth]{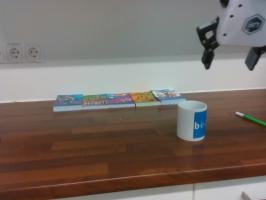}
        \end{subfigure}
        \begin{subfigure}{0.195\linewidth}
            \includegraphics[width=\linewidth]{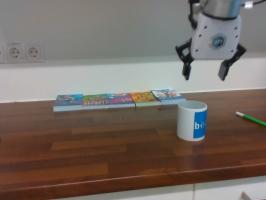}
        \end{subfigure}
        \begin{subfigure}{0.195\linewidth}
            \includegraphics[width=\linewidth]{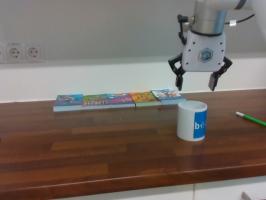}
        \end{subfigure}
        \begin{subfigure}{0.195\linewidth}
            \includegraphics[width=\linewidth]{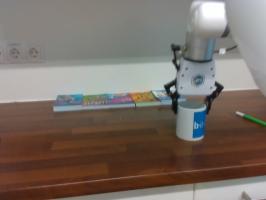}
        \end{subfigure}
        \begin{subfigure}{0.195\linewidth}
            \includegraphics[width=\linewidth]{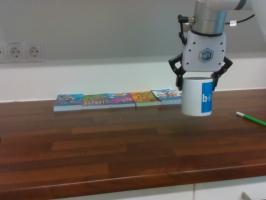}
        \end{subfigure}
        \caption{Illustration of the skill for picking up a cup}
        \label{fig:pickup-illustration}
    \end{figure*}

\begin{figure*}[t]
        \begin{subfigure}{0.195\linewidth}
            \includegraphics[width=\linewidth]{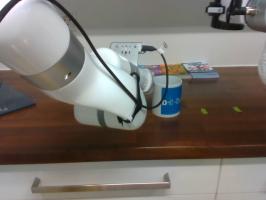}
        \end{subfigure}
        \begin{subfigure}{0.195\linewidth}
            \includegraphics[width=\linewidth]{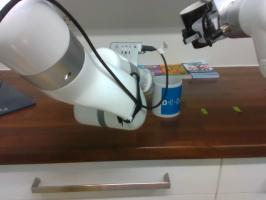}
        \end{subfigure}
        \begin{subfigure}{0.195\linewidth}
            \includegraphics[width=\linewidth]{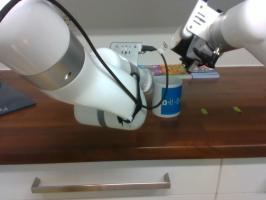}
        \end{subfigure}
        \begin{subfigure}{0.195\linewidth}
            \includegraphics[width=\linewidth]{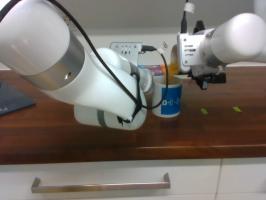}
        \end{subfigure}
        \begin{subfigure}{0.195\linewidth}
            \includegraphics[width=\linewidth]{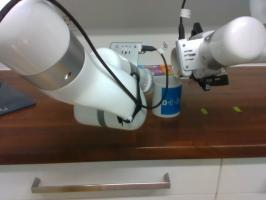}
        \end{subfigure}
        \caption{Illustration of the bimanual pouring skill (the left arm holds the cup and the right arm pours)}
        \label{fig:pour-illustration}
    \end{figure*}

    For each skill, we recorded (i) camera images (ii) proprioceptive joint data, (iii) end effector positions (with respect to the each arm's base), and (iv) delta end effector actions performed at each step.
    For the skills performed by a single manipulator, only the head camera image and the image of the respective wrist camera are collected; the images of the idle arm are excluded.
    Fig. \ref{fig:pickup-illustration} and Fig. \ref{fig:pour-illustration} illustrate successful executions of the cup pickup and pouring skills from the perspective of the robot's head camera.\footnote{It should be noted that, for safety reasons, liquid was not used in the pouring skill; we used small noodles instead.}
    For the anomalous executions, we include manually extracted annotations in the form of start and end points for the anomalies, along with textual comments on the anomalies.

    \section{IMITATION LEARNING ILLUSTRATION}
    \label{sec:imitation_learning}

\begin{figure*}[t]
        \centering
        \includegraphics[width=\linewidth]{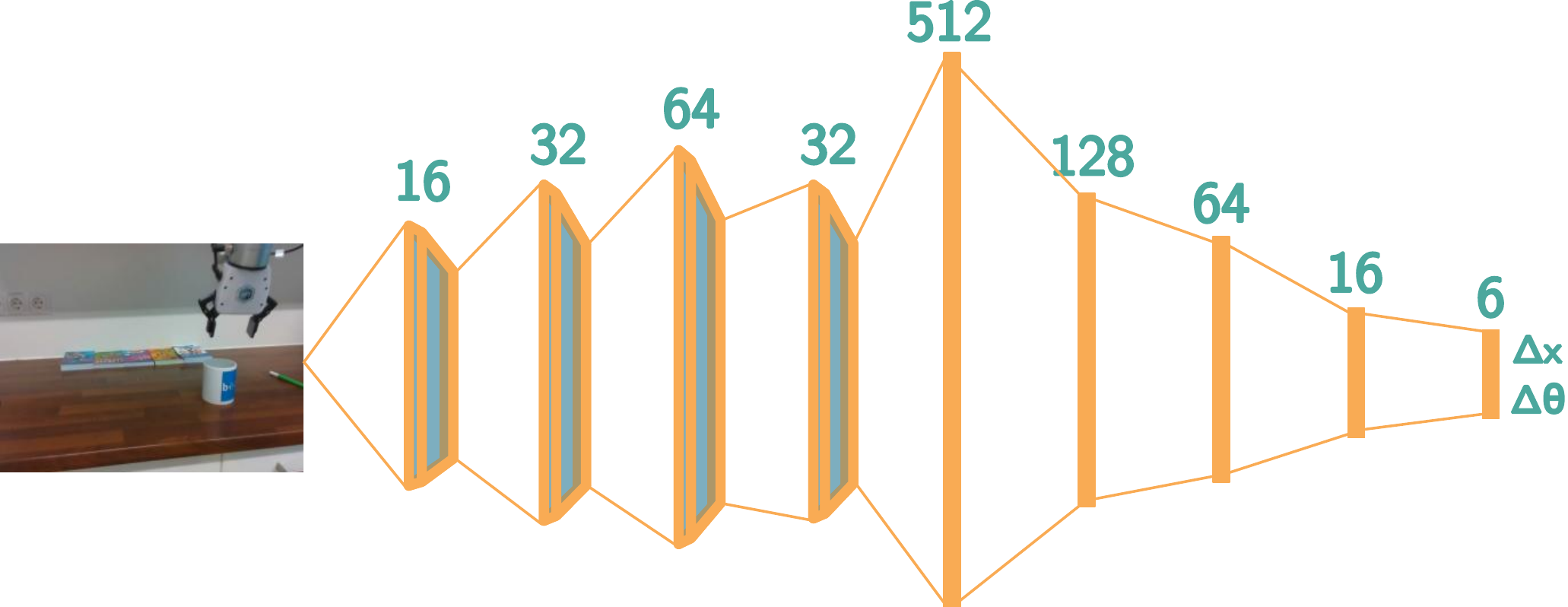}
        \caption{CNN-based policy network used to illustrate imitation learning on our data}
        \label{fig:cnn-policy}
    \end{figure*}

\begin{figure*}[t]
        \includegraphics[width=\linewidth]{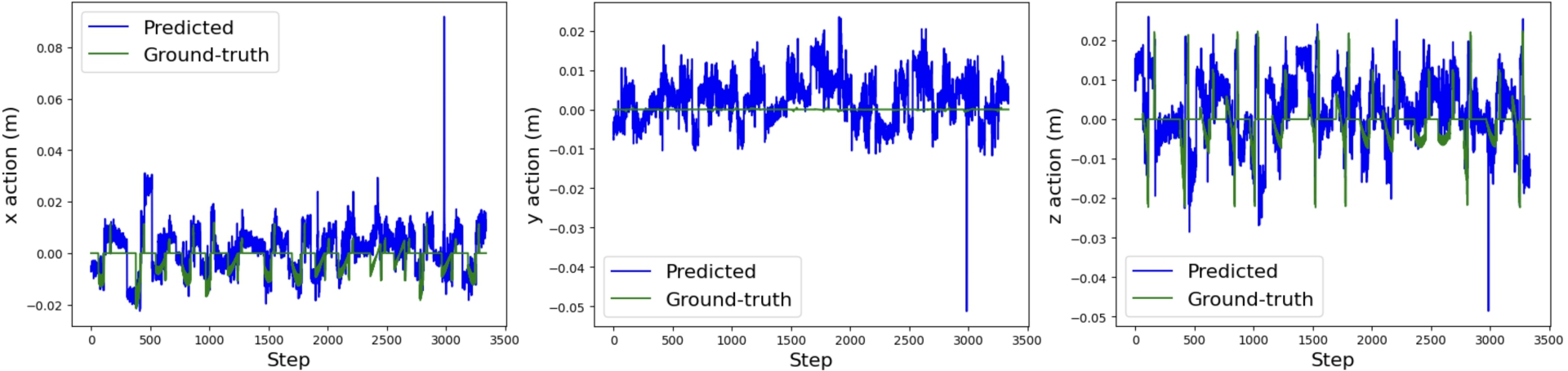}
        \caption{Predicted vs. ground-truth actions of the CNN policy}
        \label{fig:policy-predictions}
    \end{figure*}

    As COFFAIL includes both successful and anomalous executions, it can be used for different purposes, such as imitation learning, anomaly detection, and perhaps even failure recovery.
    To illustrate the use for imitation learning (using only the successful executions), we show some results on the cup pickup skill, for which we trained a convolutional neural network (CNN)-based policy as shown in Fig. \ref{fig:cnn-policy}.
    The network has multiple convolutional layers for feature extraction, such that leaky ReLU is used as activation in the feature extractor, and batch normalisation layers are included for improved training stability; in the linear layers, $\tanh$ is used as activation.
    Let $\vec{a}_t \in A$ be a ground-truth action (represented as a delta end effector pose) and $\hat{\vec{a}}_t = \pi(I_t)$, where $I_t$ is the image observation at time $t$.
    The policy $\pi$ is trained with a mean squared error (MSE) loss:
    \begin{equation*}
        \mathcal{L} = \frac{1}{T}\sum_{t=1}^{T}\lVert \vec{a}_t - \hat{\vec{a}}_t \rVert^2
    \end{equation*}
    Training of this policy was done for 30 epochs using the Adam optimiser \cite{adam_optimiser} with a learning rate of $10^{-5}$.
    Plots of the predicted and ground-truth actions are shown in Fig. \ref{fig:policy-predictions}.
    This is an example of training done in the context of one skill, but it illustrates that our dataset can be used for prototyping and investigating learning algorithms.

    \section{DISCUSSION}
    \label{sec:discussion}

    The COFFAIL dataset was collected with the intention of providing a resource for learning robot policies that take both successes and failures into account; thus, our hope is that it can be of use to researchers that aim to develop robust, failure-aware robot learning algorithms.
    In our own work, we are currently investigating how to use the dataset for learning predictive execution models as well as for extracting execution rules that can then be used for execution monitoring, but we also plan to investigate how we can perform failure diagnosis and learn recovery skills with the help of the collected data.
    Ultimately, we are interested in creating a continuous human-in-the-loop learning procedure, where diagnosis can inform operators about the type of additional data that may be needed for increasing the robustness of a particular skill.
    This may increase the usability of learning-based techniques, particularly in domains where operators may not have extensive experience with robots.


\addtolength{\textheight}{-4cm}


\bibliographystyle{IEEEtran}
\bibliography{references}

\end{document}